\def\journal{1}  

\ifcase\journal 
  \documentclass{jmlr}
\or 
  \documentclass[twoside,11pt]{article}
  \usepackage{jmlr2e}
\or 
  \documentclass{IEEEtran}
  \renewcommand{\appendix}{\appendices}
  \let\citep\cite
  \let\citet\cite
  \newcommand{\citeyearpar}[1]{\cite{#1}}
  \usepackage{cite}
  \usepackage{amsthm}
  \newtheorem{theorem}{Theorem}
  
\else
  \documentclass[10pt,a4paper]{article}
  \usepackage[square,numbers]{natbib}
  \usepackage{amsthm}
  \newtheorem{theorem}{Theorem}
\fi

\providecommand{\name}{}
\providecommand{\email}{\\email:~}
\providecommand{\addr}{~\\~}
\providecommand{\editor}[1]{}

\usepackage{amsmath}
\usepackage{amssymb}
\usepackage{amsfonts}

\usepackage{graphicx}
\usepackage{url}
\usepackage{hyperref}
\usepackage{xcolor}
\usepackage{accents} 
\usepackage[algo2e,ruled]{algorithm2e}

\usepackage{thmtools}
\declaretheorem[sibling=theorem,name=Theorem]{rtheorem}
\declaretheorem[sibling=theorem,name=Corollary]{rcorollary}

\providecommand{\theoremref}[1]{Theorem~\ref{#1}}
\providecommand{\corollaryref}[1]{Corollary~\ref{#1}}  
\providecommand{\appendixref}[1]{Appendix~\ref{#1}}

\providecommand{\sectionref}[1]{Section~\ref{#1}}
\providecommand{\figureref}[1]{Fig.~\ref{#1}}
\providecommand{\algoref}[1]{Algorithm~\ref{#1}}

\newcommand{\fnRKHS}{f} 
\newcommand{\RKHS}{\mathcal{H}} 
\newcommand{\fnSc}{s} 
\newcommand{\fnSr}{g} 
\newcommand{\fnPr}{h} 
\newcommand{\elnet}{\eta}
\newcommand{\kcoeffs}{\theta}
\newcommand{\kcsetF}{\Theta}
\newcommand{\ti}{\ell} 
\newcommand{\kk}{k} 
\newcommand{\numk}{Q} 

\newcommand{\coeffs}{\beta}
\newcommand{\iter}{m}
\newcommand{\curr}[1][\iter]{^{(#1)}}
\newcommand{\x}{x}
\newcommand{\xcurr}{w}
\newcommand{\xint}{y}
\newcommand{\xnext}{z}
\newcommand{\xsol}[1][\x]{#1^{\ast}}

\newcommand{\setA}{A}
\newcommand{\Algo}{\mathcal{A}}
\newcommand{\solSet}{\Gamma}
\newcommand{\compSet}{S}
\newcommand{\contFn}{\zeta}
\newcommand{\Mdiag}{\Lambda}
\newcommand{\objFn}{\mathcal{O}}
\newcommand{\objFnUp}{\overline{\mathcal{O}}}
\newcommand{\objFnLow}{\underline{\mathcal{O}}}
\newcommand{\alphaSVM}{\alpha}
\newcommand{\Kernel}{K}
\newcommand{\Gram}{G}
\newcommand{\normK}{u}
\newcommand{\gapSph}{d}
\newcommand{\centreSph}{c}
\newcommand{\radiusSph}{r}
\newcommand{\radiusSphcurr}{\rho}
\newcommand{\normKcurr}{w}
\newcommand{\tgtSph}{q}
\newcommand{\kcoeffssollb}{\check{\kcoeffs}}
\newcommand{\setzero}{Z}
\newcommand{\setneg}{N}

\SetKwProg{Fn}{Function}{ is}{end}%
\SetKw{KwBreak}{break}
\SetKw{KwAnd}{and}
\SetKwRepeat{Do}{do}{while}
\SetKwFunction{FSolveElNetMKL}{SolveElNetMKL}%
\SetKwFunction{FSolveElNetLP}{SolveElNetLP}%
\SetKwFunction{FSolveElNetWSR}{SolveElNetWSR}%

\newcommand{\ie}{\textit{i.e.\ }}
\newcommand{\eg}{\textit{e.g.,\ }}

\newcommand{\etal}{\textit{et al.\ }}
\newcommand{\hide}[1]{}

\newcommand{\nbd}[0]{\nobreakdash-\hspace{0pt}}
\newcommand{\Rpos}{\mathbb{R}_{\scriptscriptstyle +\!+}}
\newcommand{\Rnneg}{\mathbb{R}_{\scriptscriptstyle +}}

\newcommand{\vectgt}{\succ}
\newcommand{\vectgeq}{\succeq}
\newcommand{\vectones}{\mathbf{1}}
\newcommand{\vectzeros}{\mathbf{0}}
\newcommand{\vectbasis}{\mathbf{e}}
\newcommand{\TR}{^{\mathsf T}}
\newcommand{\scal}[2]{#1 \TR #2}
\newcommand{\grad}[1]{\nabla #1}
\newcommand{\gradi}[2]{\grad_{\!#2} #1}
\newcommand{\dee}{\mathrm{d}}
\newcommand{\minimize}{\operatornamewithlimits{minimize}}
\newcommand{\maximize}{\operatornamewithlimits{maximize}}
\newcommand{\subjto}{\operatorname{subj. to}}
\newcommand{\suml}[0]{\sum\limits}
\newcommand{\evalat}[1]{\rvert_{#1}}
\providecommand{\argmin}[1]{\operatornamewithlimits{arg\,min}}
\providecommand{\abs}[1]{\lvert#1\rvert}
\providecommand{\norm}[2][]{#1\lVert#2\rVert}
\newcommand{\numel}[1]{\lvert#1\rvert}
\newcommand{\complemInner}[2]{\settowidth{\dimen0}{$#1#2\mkern-4mu$}\accentset{\rule{\dimen0}{.6pt}}{#2}}
\newcommand{\complem}[1]{\mathpalette\complemInner{#1}}

\newcommand{\dataset}[1]{\texttt{#1}}

\usepackage{enumitem}
\newlist{lsp}{enumerate}{3}
\setlist[lsp]{label*=\arabic*.,ref=\arabic*,leftmargin=*}
\setlist[lsp,2]{ref=\thelspi.\arabic*}
\setlist[lsp,3]{ref=\thelspii.\arabic*}
\newcommand{\lspproof}{\par\nobreak{\upshape Proof:}\quad}
\newcommand{\lspqed}{Q.E.D.}

\begin{document}

\title{A simple yet efficient algorithm\\
    for multiple kernel learning
    under elastic-net constraints}

\author{\name Luca Citi \email lciti@ieee.org \\
       \addr School of Computer Science and Electronic Engineering\\
       University of Essex\\
       Colchester, CO4-3SQ, UK}

\editor{N/A}

\maketitle

\begin{abstract}
    This papers introduces an algorithm for the solution of multiple kernel
    learning (MKL) problems with elastic-net constraints on the kernel weights.
    The algorithm compares very favourably in terms of time and space complexity
    to existing approaches
    and can be implemented with simple code that does not rely on
    external libraries (except a conventional SVM solver).
\end{abstract}

\section{Introduction}

This paper presents an algorithm for the solution of
multiple kernel learning (MKL) problems with elastic-net constraints
on the kernel weights.
Please see \citet{sun_selective_2013} and \citet{yang_efficient_2011}
for a review on multiple kernel learning and its extensions.
In particular \citet{yang_efficient_2011} introduced the
generalized multiple kernel learning
(GMKL) model where the kernel weights are subject to
elastic-net constraints.

While \citet{xu_simple_2010} presents an elegant algorithm
to solve MKL problems with $L_1$\nbd norm and $L_p$\nbd norm ($p \geq 1$) constraints,
a similar algorithm is lacking in the case of
MKL under elastic-net constraints.
For example, algorithms based on the cutting plane method
\citep{yang_efficient_2011}
require large and/or commercial
libraries (\eg MOSEK).

The algorithm presented here provides an
extremely simple and efficient solution to the
elastic-net constrained MKL (GMKL) problem.
Because it can be implemented in few lines of code
and does not depend on external libraries
(except a conventional $L_2$\nbd norm SVM solver),
it has a wider applicability and can be readily
included in existing open-source
machine learning libraries.

The remainder of this paper is organised as follows.
Section~\ref{sec:MKL} introduces the elastic-net constrained
MKL problem and a solution based on a two-step block
coordinate descent method.
Two substeps of this algorithm require the solution of
more general optimisation problems, which are
therefore addressed in separate sections.
The optimization of the kernel weights relies upon the
solution of an elastic-net constrained weighted sum of reciprocals,
for which an efficient solution is presented in \sectionref{sec:elnet_wsr}.
The computation of the lower bound of the MKL cost function
(used to assess convergence of the main algorithm)
requires the solution of an elastic-net constrained linear program,
for which an efficient solution is introduced in \sectionref{sec:elnet_lp}.
Section~\ref{sec:exp_results} shows the results of a
comparison between the efficiency of the approach presented here
and that of the state-of-the-art cutting-plane method.
Finally, \sectionref{sec:conclusions} discusses the results and
draws some overall conclusions on the benefits
of the proposed approach and its potential integration
in existing machine learning libraries.

\subsection{Notation}

The symbol $\Rnneg^\numk$ denotes the set of $\numk$\nbd dimensional
vectors of nonnegative real numbers,
while $\Rpos^\numk$ the set of vectors of strictly positive real numbers.
The curled inequality symbols (\eg $\vectgt$)
represent componentwise inequality.
The symbol $\vectones_\numk$ ($\vectzeros_\numk$) denotes a $\numk \times 1$
vector of all ones (zeros) while
$\vectbasis_\kk$ is the vector
with all entries zero except the $\kk$\nbd th, which is one.
The expression $a \circ b$ computes the componentwise
product of the vectors $a$ and $b$.
The notation $\kcoeffs_\kk$ refers to
the $\kk$\nbd th component of the vector $\kcoeffs$
while $\kcoeffs\curr$ indicates the value of the vector
$\kcoeffs$ at the $m$\nbd th iteration of an iterative algorithm.
For simplicity of notation, all summations
involving $\ti$ go from 1 to $N$
(the number of training instances)
while those involving $i$ or $k$ go from 1 to $\numk$
(the number of kernels).

\section{Elastic-net constrained MKL problem}
\label{sec:MKL}

\subsection{Formulation of the generalized MKL problem}

Given a set of labelled training data $\mathcal{D} = \{x_\ti, y_\ti\}_{\ti=1}^N$
where $x_\ti \in \mathcal{X}$ and $y_\ti \in \{-1,+1\}$,
the learning problem corresponding to
a generalized MKL classifier with elastic-net constraints
\citep{yang_efficient_2011}
can be formulated as
\begin{equation}
\minimize\limits_{\substack{
\kcoeffs\in\kcsetF,
b\in\mathbb{R},\\
\{\fnRKHS_\kk\in\RKHS_\kk\}}}
\; \frac12 \suml_{\kk} \frac{\norm{\fnRKHS_\kk}^2}{\kcoeffs_\kk}
+ C \suml_\ti L \Bigl( \suml_\kk \fnRKHS_\kk(x_\ti) - b,  y_\ti \Bigr) ,
\label{eq:GMKL}
\end{equation}
where $\RKHS_\kk$ is the reproducing kernel Hilbert space (RKHS) associated with the $\kk$\nbd th kernel,
$L(\cdot)$ is the hinge loss function, and
\begin{equation}
\kcsetF = \{ \kcoeffs \in \Rnneg^\numk :
  \elnet \norm{\kcoeffs}_1 + (1-\elnet) \norm{\kcoeffs}_2^2 \leq 1 \}
  \label{eq:kcsetF}
\end{equation}
represents the elastic-net constraint on the kernel weights,
with parameter $\elnet \in [0,1]$.
When $\kcoeffs_\kk = 0$,
$\fnRKHS_\kk$ must also be equal to zero
\citep{rakotomamonjy_more_2007} and
the problem remains well-defined
(under the convention $0/0 = 0$).
Note that the minimization problem in \eqref{eq:GMKL}
is a convex optimization problem because:
a) the function to be minimized is jointly convex in its
parameters
$\kcoeffs$, $\{\fnRKHS_\kk\}$, and $b$
\citep{rakotomamonjy_more_2007};
and b) the search space is convex,
in particular the elastic-net constraint $\kcsetF$.

\subsection{Two-step block coordinate descent algorithm}

The approach taken in this manuscript for the solution of \eqref{eq:GMKL} consists of
a two-step block coordinate descent
alternating between the optimization of the SVM classifiers
and the optimization of the kernel weights.
The procedure, which is reported in \algoref{algo:outer},
iterates until a stopping condition is met
(see \sectionref{sec:lower_bound_MKL}).

  \begin{small}
  \begin{algorithm2e} 
  \Fn{\FSolveElNetMKL{$\Gram_1,\dots,\Gram_\numk,y$}}{
    \tcp{Initialization}
    $\kcoeffs \leftarrow \vectones_\numk / \fnSc(\vectones_\numk)$\;
    \For{$\iter \leftarrow 1$ \KwTo maximum number of iterations}{
      \tcp{Step 1: optimization of the SVM classifiers}
      build composite Gram matrix: $G \leftarrow \sum_{\kk} \kcoeffs_\kk G_\kk$\;
      solve std SVM with $G,y$ to get optimal dual coeffs and bias: $\alphaSVM,b$\;
      \tcp{Step 1.5: check convergence}
      \lFor{$\kk \leftarrow 1$ \KwTo $\numk$}{
        $\normK_\kk \leftarrow
            [(\alphaSVM\circ y)\TR \, G_\kk \, (\alphaSVM\circ y)] $
      }
      compute objective function \eqref{eq:GMKL} from dual form of SVM:
          $\objFnUp \leftarrow \scal{\vectones_{\numk}}{\alphaSVM}
                                - \scal{\normK}{\kcoeffs} $\;
      solve elastic-net constrained LP: $\kcoeffssollb \leftarrow$
          \FSolveElNetLP{$\normK$} \;
      compute lower bound of \eqref{eq:GMKL}:
          $\objFnLow \leftarrow \scal{\vectones_{\numk}}{\alphaSVM}
                                - \scal{\normK}{\kcoeffssollb} $\;
      \lIf{$\objFnUp/\objFnLow - 1 < \epsilon_{\mathrm{MKL}}$}{\KwBreak}
      \tcp{Step 2: optimization of the kernel weights}
      compute $\norm{\fnRKHS_\kk}^2$:
          $\coeffs \leftarrow \kcoeffs \circ \kcoeffs \circ \normK$\;
      solve elastic-net constr.\ weighted sum of recipr.: $\kcoeffs \leftarrow$
          \FSolveElNetWSR{$\coeffs$, $\kcoeffs$}\;
    }
    \KwRet{$\kcoeffs,\alphaSVM,b$}\;
  }
  \caption{Solve elastic-net constrained MKL.}\label{algo:outer}
  \end{algorithm2e}
  \end{small}

At iteration $\iter$, the first step minimizes problem \eqref{eq:GMKL}
with respect to $\{\fnRKHS_\kk\}$ and $b$ for fixed
values of the kernel weights $\kcoeffs\curr$. As already noted by others
\citep{rakotomamonjy_more_2007,xu_simple_2010,yang_efficient_2011},
this problem is equivalent to the standard SVM problem with a
composite kernel
$\Kernel(\cdot, \cdot) = \sum_{\kk} \kcoeffs\curr_\kk \Kernel_\kk(\cdot, \cdot)$.
Given the stack of Gram matrices $\Gram_{\kk,\ti,\ti'}$,
where $\Gram_{\kk,\ti,\ti'} = \Kernel_\kk(x_\ti, x_{\ti'})$,
existing SVM solvers can efficiently solve
the composite SVM problem with Gram matrix
$\Gram\curr = \sum_{\kk} \kcoeffs\curr_\kk \Gram_\kk$ and return
the optimal bias $b\curr$ and
vector of dual coefficients $\alphaSVM\curr$.

The second step consists in minimizing \eqref{eq:GMKL}
for $\kcoeffs \in \kcsetF$ while keeping $b$ and $\{\fnRKHS_\kk\}$ (or equivalently the dual coefficients)
constant.
Since the only term that depends on $\kcoeffs$ is the regularizer,
we can define
\begin{equation}
  \coeffs\curr_\kk = \norm{\fnRKHS\curr_\kk}^2 = (\kcoeffs\curr_\kk)^2 \, [(\alphaSVM\curr\circ y)\TR \, G\curr_\kk \, (\alphaSVM\curr\circ y)], \quad \forall \kk,
\end{equation}
and attack this sub-problem as an instance of the more general problem of
minimizing a weighted sum of reciprocals bound to
elastic-net constraints:
\begin{equation}
\kcoeffs\curr[\iter+1] = \argmin\limits_{\kcoeffs \in \kcsetF} \; \suml_\kk \frac{\coeffs\curr_\kk}{\kcoeffs_\kk}.
\label{eq:WSR_for_MKL}
\end{equation}
Assuming positive definite kernels
and excluding degenerate cases causing $\alphaSVM\curr = \vectzeros_\numk$
(\eg all examples belonging to the same class),
we have that $\coeffs\curr \vectgt 0$ as long as $\kcoeffs\curr \vectgt 0$.
Because \eqref{eq:WSR_for_MKL} diverges to $+\infty$
when any $\kcoeffs_\kk$ approaches zero, the
minimization of \eqref{eq:WSR_for_MKL} will always produce
$\kcoeffs\curr[\iter+1] \vectgt 0$ as long as
$\kcoeffs\curr \vectgt 0$, \ie ultimately provided that
the initial point
$\kcoeffs\curr[0] \vectgt 0$.

In the special case $\elnet = 1$,
the elastic-net constraint reduces to a lasso constraint and the
problem \eqref{eq:WSR_for_MKL}
has a straightforward closed-form solution \citep{xu_simple_2010}.
In this manuscript, a novel, simple and efficient algorithm for the solution of
this optimization problem in the general case $\elnet \in [0,1]$
is presented.
Since the proposed solution to this sub-problem
represents the novelty and main contribution
of this paper, \sectionref{sec:elnet_wsr} will be
entirely devoted to explaining this algorithm in detail.

\subsection{Lower bound and stopping condition}
\label{sec:lower_bound_MKL}

Establishing a lower bound on the optimal value
of the cost function 
\eqref{eq:GMKL}
provides a non-heuristic stopping criterion for the
outer iterative algorithm, \ie the
two-step block coordinate descent method.
Following \citep{yang_efficient_2011}, the lower bound is found
as the minimum over $\kcoeffs$ of
the dual form of \eqref{eq:GMKL}:
\begin{equation}
\minimize\limits_{\kcoeffs\in\kcsetF}
    \vectones_\numk \TR \, \alphaSVM - 
    \frac{1}{2}\,
    (\alphaSVM\circ y)\TR
    \left( \suml_\kk \kcoeffs_\kk G_\kk \right)
    \, (\alphaSVM\circ y),
\label{eq:dual}
\end{equation}
where $\alphaSVM$ is the
vector of dual coefficients of the composite SVM problem.
In Yang \etal \citeyearpar{yang_efficient_2011}, this bound
is obtained as part of the cutting-plane method used
for the optimization of the kernel weights.
The method proposed here takes a radically different approach as
it finds the point $\kcoeffssollb$ where the minimum of
\eqref{eq:dual} is attained as the solution of
the elastic-net constrained linear program:
\begin{equation}
\maximize\limits_{\kcoeffs\in\kcsetF}
\; \scal{\normK}{\kcoeffs},
\label{eq:elnet_LP_for_MKL}
\end{equation}
where
\begin{equation}
    \normK_\kk =
    (\alphaSVM\circ y)\TR G_\kk \, (\alphaSVM\circ y), \quad \forall \kk.
\end{equation}
A novel, simple and efficient algorithm for the solution of \eqref{eq:elnet_LP_for_MKL}
is provided in \sectionref{sec:elnet_lp}.

At each iteration, problem \eqref{eq:elnet_LP_for_MKL} is solved for the current iterates $\alphaSVM\curr$ and $\kcoeffs\curr$. The current value of the objective function and of the lower bound are simply computed as
\begin{align}
  \objFnUp\curr &= \vectones_\numk \TR \, \alphaSVM\curr -
                   \frac{1}{2}\,\scal{\bigl(\normK\curr\bigr)\!}{\kcoeffs\curr}
   &
  \objFnLow\curr &= \vectones_\numk \TR \, \alphaSVM\curr -
                    \frac{1}{2}\,\scal{\bigl(\normK\curr\bigr)\!}{\kcoeffssollb\curr}.
\end{align}
The two-step block coordinate descent algorithm
terminates when an iterate with relative gap
$\objFnUp\curr\!/\objFnLow\curr - 1 < \epsilon_{\mathrm{MKL}}$ is produced,
which guarantees that
the current value of the objective function $\objFnUp\curr$
is at most $\epsilon_{\mathrm{MKL}} \, \objFn\curr[\infty]$
away from the optimal value $\objFn\curr[\infty]$.

\section{Elastic-net constrained weighted sum of reciprocals}
\label{sec:elnet_wsr}

This whole section abstracts from the original MKL learning problem
and focuses on the solution
of the following optimization problem:
\begin{equation}
\begin{split}
\minimize \quad & \suml_\kk \frac{\coeffs_\kk}{\kcoeffs_\kk}\\
\subjto \quad
  & \elnet \norm{\kcoeffs}_1 + (1-\elnet) \norm{\kcoeffs}_2^2 \leq 1,\\
  & \kcoeffs \vectgeq 0
\end{split}
\label{eq:WSR}
\end{equation}
with $\coeffs \vectgt 0$ and $\elnet \in [0,1]$.
As mentioned before, a solution to this problem must lie
in the strictly positive orthant $\kcoeffs \vectgt 0$.
Furthermore, the solution must be attained at a point where
the elastic-net constraint is tight, \ie
$\elnet \norm{\kcoeffs}_1 + (1-\elnet) \norm{\kcoeffs}_2^2 = 1$.
Aiming for a contradiction, let us assume that $\kcoeffs$
minimizes \eqref{eq:WSR} with
$\elnet \norm{\kcoeffs}_1 + (1-\elnet) \norm{\kcoeffs}_2^2 = 1 - \epsilon$,
$0 < \epsilon \leq 1$.
The point
$\kcoeffs' = (1+\frac{\epsilon}{3})\,\kcoeffs$ clearly decreases the
cost function while still satisfying the elastic-net constraint:
$\elnet \norm{\kcoeffs'}_1 + (1-\elnet) \norm{\kcoeffs'}_2^2 <
(1+\frac{\epsilon}{3}) \elnet \norm{\kcoeffs}_1 +
  (1+\frac{7}{9}\epsilon) (1-\elnet) \norm{\kcoeffs}_2^2
< (1+\epsilon)\,(1-\epsilon) < 1$.
This contradicts the original assumption that $\kcoeffs$
was a minimum for \eqref{eq:WSR}.
Therefore, we can search for the solution to \eqref{eq:WSR}
among the points in $\Rpos^{\numk}$ for which the
elastic-net constraint holds with equality.
Please notice that in the remainder of this section $x$ and $y$ simply denote
vectors in $\Rpos^{\numk}$ (rather than training instances and labels
like in the previous sections).

\subsection{Re-scaled objective function}
\label{sec:rescaled}

As a preliminary step in attacking the problem \eqref{eq:WSR},
we introduce an equivalent optimization problem.
It is easy to verify that the norm
\begin{equation}
\fnSc(x) = \frac{\elnet}{2} \, \norm{\x}_1 + \sqrt{\frac{\elnet^2}{4}\, \norm{\x}_1^2 + (1-\elnet) \, \norm{\x}_2^2}  \label{eq:fnSc}
\end{equation}
verifies
$\elnet \norm[\bigl]{\frac{\x}{\fnSc(\x)}\bigr}_1 + (1-\elnet) \norm[\bigl]{\frac{\x}{\fnSc(\x)}\bigr}_2^2 = 1$,
$\forall \x \in \mathbb{R}^\numk \setminus \{0\}$.
As a result, the change of variable $\kcoeffs = \x / \fnSc(\x)$,
transforms the original problem \eqref{eq:WSR}
into the following equivalent one:
\begin{equation}
\minimize\limits_{\x \in \Rpos^\numk} \quad \fnPr(\x)
= \fnSc(\x)\,\fnSr(\x)
\label{eq:equiv_problem}
\end{equation}
where
\begin{equation}
\fnSr(\x) = \suml_i \frac{\coeffs_i}{\x_i}. \label{eq:fnSr}
\end{equation}
This new optimization problem implicitly accounts for
the elastic-net constraint
by means of the rescaling function $\fnSc$
which re-normalizes any $\x \in \Rpos^Q$ such that
the vector $\kcoeffs = \x / \fnSc(\x)$ satisfies the
elastic-net constraint with equality.
Our new task is therefore to find a global minimum of
$\fnPr$ in the positive orthant.
Although $\fnPr$ is not a convex function,
we can prove a
weaker result --- pseudoconvexity ---
which is still very useful in practice
because critical points of pseudoconvex
functions are also global minima
\citep[theorem 3.2.5]{cambini_generalized_2008}.
In order to show that $\fnPr$ is pseudoconvex,
the following theorem and its corollary are introduced (proofs in \appendixref{sec:proof_pseudoconvexity}).
\begin{rtheorem}[name=,restate=pseudoconvexity]
\label{th:pseudoconvexity}%
Let $\setA \subseteq \mathbb{R}^n$ be an open convex cone and
$\fnSr,\fnSc: \setA \rightarrow \Rpos$
be differentiable convex functions such that
$\fnSc(c\x) = c \fnSc(\x)$
and
$\fnSr(c\x) = \fnSr(\x)/c$
for all $c \in \Rpos$ and $\x \in \setA$.
Their pointwise product
$\fnPr(\x) = \fnSc(\x)\,\fnSr(\x)$
is a pseudoconvex function in $\setA$.
\end{rtheorem}
\begin{rcorollary}[name=,restate=pseudoconvexitycoroll]
\label{th:pseudoconvexitycoroll}%
Under the conditions of \theoremref{th:pseudoconvexity},
all points $\x = c\xsol$ --- with $c \in \Rpos$
and $\xsol$ satisfying
$\grad{\fnSc(\xsol)} = -\grad{\fnSr(\xsol)}$ ---
are global minima for the function $\fnPr$,
where it takes value $\fnPr(c\xsol)=\fnSc^2(\xsol)=\fnSr^2(\xsol)$.
If at least one of $\fnSc$ or $\fnSr$ is strictly convex, then
$\xsol$ is unique.
\end{rcorollary}

The functions $\fnSc$ and $\fnSr$ defined in \eqref{eq:fnSr} and \eqref{eq:fnSc}
satisfy the requirements for \theoremref{th:pseudoconvexity}
because they are positive-valued differentiable functions
in the positive orthant $\Rpos^{\numk}$ (which is an open convex cone)
and they can be shown to be convex through some simple calculus.
As a result of \theoremref{th:pseudoconvexity},
$\fnPr$ is pseudoconvex function in $\Rpos^{\numk}$.
Additionally, the strict convexity of $\fnSr$ guarantees the uniqueness
of $\xsol$ defined in \corollaryref{th:pseudoconvexitycoroll}.


\subsection{Iterative minimization algorithm}
\label{sec:algorithm}

The problem \eqref{eq:equiv_problem} can be minimized
using the following novel iterative algorithm.\footnote{
Please notice that the superscript $\iter$ now refers
to the current iteration within the algorithm for the solution of
\eqref{eq:WSR} and is completely unrelated to the current
iteration in the outer two-step block coordinate descent
algorithm for the solution of the original MKL problem.}
Given the current iterate $\x\curr$,
the next iterate $\x\curr[\iter+1]$ is generated as:
\begin{subequations}
\label{eq:iterate}
\begin{equation}
\label{eq:iterate_x}
    \x\curr[\iter+1]_i = \sqrt{\frac{\coeffs_i}{q_i\curr}}
\end{equation}
where
\begin{equation}
\label{eq:iterate_q}
    q_i\curr = \gradi{\fnSc(\x\curr)}{i} =
      \left. \frac{\dee \fnSc(\x)}{\dee \x_i} \right\evalat{\x=\x\curr}
    .
\end{equation}
\end{subequations}
The algorithm is iterated until a stopping condition is met,
at which point the last iterate $\hat \x$ is re-scaled
to obtain the solution to the problem \eqref{eq:WSR}
as $\hat \kcoeffs = \hat \x / \fnSc(\hat \x)$.
The pseudocode of the full algorithm ---
including the stopping condition
that will be described in the following ---
is reported in \algoref{algo:wsr}.

  \begin{small}
  \begin{algorithm2e} 
  \Fn{\FSolveElNetWSR{$\coeffs$, $\kcoeffs\curr[0]$}}{
    \tcp{Initialization}
    $\x \leftarrow \kcoeffs\curr[0]$\;
    \For{$\iter \leftarrow 1$ \KwTo maximum number of iterations}{
      \tcp{Compute cost}
      $n_1 \leftarrow \sum_{\kk} \x_\kk$\;
      $r \leftarrow \sqrt{(\tfrac{\elnet}{2})^2 n_1^2  + (1 - \elnet) \sum_{\kk} \x_\kk^2}$\;
      $\fnSc \leftarrow  \tfrac{\elnet}{2} n_1 + r$\;
      $\fnSr \leftarrow  \sum_{\kk} \coeffs_\kk / \x_\kk$\;
      \tcp{Check convergence}
      \lIf{$\iter > 1$ \KwAnd $\fnSc/\fnSr - 1 < \epsilon_{\mathrm{wsr}}$}{\KwBreak}
      \tcp{Update iterate}
      $q \leftarrow \tfrac{\elnet}{2} + [(\tfrac{\elnet}{2})^2 n_1 + (1 - \elnet)\,\x] / r$\;
      \lFor{$\kk \leftarrow 1$ \KwTo $\numk$}{
        $\x_\kk \leftarrow \sqrt{\coeffs_\kk / q_\kk} $
      }
    }
    \KwRet{$\x/s$}\;
  }
  \caption{Solve elastic-net constrained weighted sum of reciprocals.}\label{algo:wsr}
  \end{algorithm2e}
  \end{small}

While a full proof of the convergence of the algorithm is provided in \sectionref{sec:ext_tan_proof},
the intuition behind it is sketched here.
For ease of notation,
we will hereafter drop the iteration superscript
and refer to the current iterate as
$\xcurr \triangleq \x\curr$
and to the next one as
$\xnext \triangleq \x\curr[\iter+1]$.
The new iterate $\xnext$ generated from \eqref{eq:iterate}
can be interpreted as the solution to the 
problem:
\begin{equation}
\begin{split}
\minimize\limits_{\x \in \Rpos^\numk} \quad & \suml_i \frac{\coeffs_i}{\x_i}\\
\subjto \quad & \scal{q}{\x} = p,
\label{eq:tan-min-constr}
\end{split}
\end{equation}
where $p = \sum_i \sqrt{\coeffs_i q_i}$.
In other words, the new iterate is generated by
minimizing the function $\fnSr$ on a hyperplane
which is perpendicular to the gradient of $\fnSc$ at $\xcurr$.
The specific choice of the offset constant, \ie $p$ in \eqref{eq:tan-min-constr},
has an interesting geometrical interpretation.
%
Because the functions $\fnSc$ and $\fnSr$ satisfy the requirements for \theoremref{th:pseudoconvexity},
for any positive $c$ the point $\xint = c\xcurr$ is such that
$\fnSc(\xint)\fnSr(\xint) = \fnSc(\xcurr)\fnSr(\xcurr)$
and also that
$p = \scal{q}{\x} = \scal{\grad{\fnSc(\xint)}}{\x} \leq \fnSc(\x)$ $\forall \x$
(see \theoremref{th:boundsfnSc} below).
Choosing $c$ such that $\fnSc(\xint) = c \fnSc(\xcurr) = p$ and
substituting \eqref{eq:iterate} in \eqref{eq:fnSr}, it is easy to
show that the hyperplane $\scal{q}{\x} = p$ has the following properties:
\begin{subequations}
\label{eq:ext_tan}
\begin{alignat}{2}
  q &= \grad{\fnSc(\xint)} = -\grad{\fnSr(\xnext)},\label{eq:ext_tan_grad}\\ 
  p &= \fnSc(\xint)  \leq \fnSc(\x) &\quad& \text{and}\\
  p &= \fnSr(\xnext) \leq \fnSr(\x) && \forall \x:\scal{q}{\x} = p. \label{eq:ext_tan_Sr}
\end{alignat}
\end{subequations}
In other words, this hyperplane is externally
tangent to the level sets of $\fnSc$ and  $\fnSr$
of the same value, $p$.
Asymptotically, the algorithm finds the hyperplane
that is tangent to the two level sets at the same point.
Although there is no guarantee that each
step decreases both $\fnSr$ and $\fnSc$,
the next section will show that their product
$\fnPr$ decreases monotonically at each step
and that the algorithm effectively convergences towards the solution.

\subsection{Convergence analysis}
\label{sec:ext_tan_proof}

A fixed point for the iterative map \eqref{eq:iterate}
is the point $\xsol$ satisfying the conditions of \corollaryref{th:pseudoconvexitycoroll},
since substituting
$\xsol[q_i] = \gradi{\fnSc(\xsol)}{i} = -\gradi{\fnSr(\xsol)}{i} = \coeffs_i/(\xsol_i)^2$ in the iterate update \eqref{eq:iterate_x} makes it an identity.
By \corollaryref{th:pseudoconvexitycoroll}, this fixed point
is a global minimum for $\fnPr$ and, therefore,
a solution for \eqref{eq:equiv_problem}.

To show that the algorithm \eqref{eq:iterate} can be used to solve
\eqref{eq:equiv_problem}, it remains to be proven that the iterative map \eqref{eq:iterate}
converges to its fixed point $\xsol$ for all starting points
$\x\curr[0] \in \Rpos^{\numk}$.
To do so, we will make use of convergence results
of descent algorithms
\citep{zangwill_nonlinear_1969,meyer_sufficient_1976,bertsekas_nonlinear_1999,luenberger_linear_2008}
and in particular of Zangwill's
Global Convergence Theorem \citep[p. 205]{luenberger_linear_2008},
restated here for convenience.
\begin{rtheorem}[name=Global Convergence Theorem,restate=globalconverg]
\label{th:globalconverg}%
Let $\Algo$ be an algorithm on $\setA$, and suppose that,
given $\x\curr[0]$, the sequence $\{\x\curr\}_{m=0}^{\infty}$
is generated satisfying $\x\curr[\iter+1] \in \Algo(\x\curr)$.
Let a solution set $\solSet \subset \setA$ be given, and suppose:
\begin{enumerate}[label=\arabic*.,ref=\arabic*]
  \item\label{th:globalconverg_compact}
    all points $\x\curr$ are contained in a compact set $\compSet \subset \setA$,
  \item\label{th:globalconverg_decr}
  there is a continuous function $\contFn$ on $\setA$ such that:
  \begin{enumerate}[label=(\alph*),ref=\theenumi.(\alph*)]
    \item if $\x \notin \solSet$, then $\contFn(\xnext) < \contFn(\x)$ for all $\xnext \in \Algo(\x)$, \label{th:globalconverg_decr_lt}
    \item if $\x \in \solSet$, then $\contFn(\xnext) \leq \contFn(\x)$ for all $\xnext \in \Algo(\x)$, \label{th:globalconverg_decr_leq}
  \end{enumerate}
  \item\label{th:globalconverg_closedmap}
  the mapping $\Algo$ is closed at points outside $\solSet$.
\end{enumerate}
Then the limit of any convergent subsequence of $\{\x\curr\}$ is a solution.
\end{rtheorem}
The following will show that \theoremref{th:globalconverg} applies to
the mapping $\{\x\curr[\iter+1]\} = \Algo(\x\curr)$
corresponding to \eqref{eq:iterate}.
This mapping is defined in $\setA = \Rpos^{\numk}$ and
has solution set $\solSet = \{\xsol\}$.

Since $\fnSc$ is a differentiable convex function in the open convex set $\Rpos^{\numk}$,
it is actually continuously differentiable in $\Rpos^{\numk}$
\cite[Corollary 25.5.1]{rockafellar_convex_1970}.
The specific choice of $\fnSc$ in \eqref{eq:fnSc} is such that
$q_i\curr$ is also strictly positive and, therefore,
the iteration \eqref{eq:iterate} defines a continuous function (point-to-point mapping)
from $\x\curr$ to
$\x\curr[\iter+1]$.
Since for a point-to-point mapping
continuity implies closedness \citep[p. 206]{luenberger_linear_2008},
the third condition of Zangwill's theorem is satisfied.

As a first step towards verifying the second condition,
the following theorem is introduced (proof provided in \appendixref{sec:proof_boundsfnSc}).
\begin{rtheorem}[name=,restate=boundsfnSc]
\label{th:boundsfnSc}%
Given a norm $s: \mathbb{R}^n \rightarrow \Rnneg$
of the form
$\fnSc(\x) = d_0 \norm{\x}_1 + \sqrt{d_1 \norm{\x}_1^2 + d_2 \norm{\x}_2^2}$ with $d_0, d_1, d_2 \geq 0$, the following property holds:
\begin{equation}
  \scal{\grad{\fnSc(\xint)}}{\x} \leq
  \fnSc(\x) \leq \sqrt{\x\TR \Mdiag_{\xint} \, \x}
    \quad \forall \x, \xint \in \mathbb{R}^n,
\end{equation}
where $\Mdiag_{\xint}$ is a diagonal matrix whose $i$\nbd th diagonal element is
$\fnSc(\xint) \gradi{\fnSc(\xint)}{i} / \xint_i$.
\end{rtheorem}
We can now write the following chain of inequalities
showing that $\fnPr$ is non-increasing at each step:
\begin{equation}
\label{eq:fnPr-nonincreasing}
\fnPr(\x\curr[\iter+1])
\equiv \fnPr(\xnext)
\leq \fnSc^2(\xnext) \leq \xnext\TR \Mdiag_{\xint} \, \xnext
= \suml_i \sqrt{\frac{\coeffs_i}{q_i}} \,
       \fnSc(\xint) \frac{q_i}{\xint_i} \,
       \sqrt{\frac{\coeffs_i}{q_i}}
= \fnPr(\xint)
= \fnPr(\xcurr)
\equiv \fnPr(\x\curr),
\end{equation}
where the first inequality follows from \eqref{eq:ext_tan}
while the second one from \theoremref{th:boundsfnSc}.
Unfortunately, the fact that $\fnPr$ is constant along
rays out of the origin makes it unsuitable as function $\contFn$
for \theoremref{th:globalconverg}
(the strict inequality in condition \ref{th:globalconverg_decr_lt} is violated
for points $\x = c \xsol$ with $c \in \Rpos$).
Instead, we consider the function
\begin{equation}
\label{eq:contFn}
\contFn(\x) = 2 \, \fnPr(\x) + \left[\fnSc(\x) - \fnSr(\x)\right]^2
  = \fnSr^2(\x) + \fnSc^2(\x) ,
\end{equation}
for which the following inequality can be readily obtained from
\eqref{eq:ext_tan}, \eqref{eq:fnPr-nonincreasing}, and \eqref{eq:contFn}:
\begin{equation}
\label{eq:contFn_ineq}
\contFn(\xnext) = \fnSr^2(\xnext) + \fnSc^2(\xnext) \leq
2 \, \fnSc^2(\xnext) \leq 2 \, \fnPr(\xcurr) \leq
\contFn(\xcurr).
\end{equation}
Importantly, as prescribed by \ref{th:globalconverg_decr_lt},
the expression \eqref{eq:contFn_ineq} holds with equality only if
the starting point of the iteration ($\xcurr$ in our case)
is in the solution set $\solSet$.
This can be shown by first noticing that
$\contFn(\xnext) = \contFn(\xcurr)$ implies
$\fnSr(\xnext) = \fnSc(\xnext) = \fnSr(\xcurr) = \fnSc(\xcurr)$.
From the definition of $\xint$, we see that
$\fnSc(\xcurr) = \fnSr(\xnext) \; \Rightarrow \; \xint \equiv \xcurr$.
Since the restriction of $\fnSr$ along $\scal{q}{\x} = p$ is strictly
convex, the inequality in \eqref{eq:ext_tan_Sr} holds as equality
only at the minimum, \ie
$\fnSr(\xnext) = \fnSr(\xint) \; \Rightarrow \; \xnext \equiv \xint$.
Putting these together, we obtain that $\xnext \equiv \xcurr$,
which substituted in \eqref{eq:ext_tan_grad} finally yields
$\grad{\fnSc(\xcurr)} = -\grad{\fnSr(\xcurr)}$,
the condition defining the fixed point $\xsol$.
This proves that the second condition of Zangwill's theorem
is also satisfied.

Through some simple algebra, it is easy to show that
$\gradi{\fnSc(\x)}{i} \leq 1$ $\forall \x,i$.
This, together with \eqref{eq:fnSc} and \eqref{eq:fnPr-nonincreasing},
leads to
$\sqrt{\coeffs_i} \leq \x_i\curr[\iter+1]
\leq \fnSc(\x\curr[\iter+1]) \leq \sqrt{\fnPr(\x\curr)} \leq \sqrt{\fnPr(\x\curr[0])}$.
As a result, all the points of the sequence
(with the immaterial possible exception of $\x\curr[0]$)
are contained in $[\, \min_i \sqrt{\coeffs_i},\, \sqrt{\fnPr(\x\curr[0])} \,]^{\numk}$, which is a closed and bounded subset of
$\Rpos^{\numk}$, as prescribed by the first condition of the theorem.

In conclusion, we have proven that the algorithm defined
by the iteration \eqref{eq:iterate} satisfies the
conditions of Zangwill's theorem.
Also, because the solution set $\solSet$ consists of
a single point $\xsol$, the sequence $\{\x\curr\}$
converges to $\xsol$ \citep[p. 206]{luenberger_linear_2008}.

\subsection{Stopping condition}
\label{sec:stopping}

We now establish a lower bound on the optimal value
of $\fnPr$, which will be used to
provide a non-heuristic stopping criterion for the
iterative algorithm in \eqref{eq:iterate}.
Given the solution $\xsol$ and the new iterate $\x\curr[\iter+1]$
obtained as described in \sectionref{sec:algorithm}, we observe
that, since $q$ and $\xsol$ lie in the (strictly)
positive orthant, there always exists $c \in \Rpos$ such a that
$\scal{q}{(c\xsol)} = p$, with $p$ as in \sectionref{sec:algorithm}.
Therefore, \eqref{eq:ext_tan_Sr} implies
$p \leq \fnSr(c\xsol)$ and \theoremref{th:boundsfnSc}
yields $p = \scal{q}{(c\xsol)} \leq \fnSc(c\xsol)$.
Combining these two inequalities gives
$p^2 \leq \fnSr(c\xsol) \, \fnSc(c\xsol)$,
which can be rewritten as $\fnSr^2(\x\curr[\iter+1]) \leq \fnPr(\xsol)$
where the equality only holds at the solution $\xsol$.
As a result, $\fnPr(\x\curr[\iter+1]) - \fnSr^2(\x\curr[\iter+1])$
bounds how suboptimal the iterate is, even
without knowing the exact value of $\fnPr(\xsol)$.
The following stopping condition guarantees
a predefined relative accuracy $\epsilon_{\mathrm{wsr}} > 0$:
\begin{equation}
    \frac{\fnPr(\x\curr[\iter+1]) - \fnSr^2(\x\curr[\iter+1])}{\fnSr^2(\x\curr[\iter+1])}
    = \frac{\fnSc(\x\curr[\iter+1])}{\fnSr(\x\curr[\iter+1])} - 1
    \leq \epsilon_{\mathrm{wsr}}.
    \label{eq:terminate}
\end{equation}
The algorithm terminates after
an $\epsilon_{\mathrm{wsr}}$\nbd suboptimal iterate is produced, \ie
when \eqref{eq:terminate} is satisfied,
which guarantees that
$\fnPr(\x\curr[\iter+1]) - \fnPr(\xsol) \leq \epsilon_{\mathrm{wsr}} \, \fnPr(\xsol)$.

\subsection{Alternative approaches}
\label{sec:wsr-alternative}

This section presents a brief overview of alternative approaches
that were devised by the author in the process of creating and improving
the main method presented above.
They are reported here because they may be advantageous in
specific situations and for some values of the parameters.

An approach to minimizing \eqref{eq:WSR},
which works particularly well when $\elnet$ is small,
is by using the alternative update:
\begin{equation}
\label{eq:iterate_x_alt}
    \x\curr[\iter+1]_i = \left(\frac{\x\curr_i\coeffs_i}{q_i\curr}\right)^{\!\frac13}
\end{equation}
instead of \eqref{eq:iterate_x}.
For an appropriate choice of $p'>0$,
this iterate is the solution to the problem:
\begin{equation}
\begin{split}
\minimize\limits_{\x \in \Rpos^\numk} \quad & \suml_i \frac{\coeffs_i}{\x_i}\\
\subjto \quad & \x\TR \Mdiag_{\x\curr} \, \x = p', 
\label{eq:tan-min-constr-sq}
\end{split}
\end{equation}
which is an analogous of \eqref{eq:tan-min-constr}
using a quadratic constraint instead of a linear one.
The iterative map defined by \eqref{eq:iterate_x_alt}
has the same fixed point as the map \eqref{eq:iterate_x} and
a convergence proof can be obtained using arguments similar
to those in \sectionref{sec:ext_tan_proof}.
In simulations, the convergence rate of the update rule \eqref{eq:iterate_x_alt},
appears to be marginally better than \eqref{eq:iterate_x} for small
values of $\elnet$ (less than approximately $0.25$) and
significantly worse otherwise.
For this reason, it may be advantageous to use \eqref{eq:iterate_x}
when $\elnet \geq 0.25$ and alternate between \eqref{eq:iterate_x_alt}
and \eqref{eq:iterate_x} when $\elnet < 0.25$.

An algorithm for the solution of the problem \eqref{eq:WSR}
using a majorization-minimization (MM) procedure was presented
in \citep{citi_elastic-net-MM_2015}.
Briefly, the algorithm is similar to coordinate descent
but at each step ---
instead of performing a full line search to minimize
$\fnPr$ as a function of one of the optimization variables ---
it reduces it by minimizing a carefully designed surrogate function,
called a majorizer, which can be solved in closed form.
The number of iterations required to obtain a given accuracy
is comparable to that of \algoref{algo:wsr} but each
iteration requires roughly four times as many flops.

\section{Elastic-net constrained linear program}
\label{sec:elnet_lp}

This section introduces an efficient algorithm for finding the
solution $\kcoeffssollb$ of the elastic-net constrained linear program:
\begin{equation}
\begin{split}
\maximize \quad & \scal{\normK}{\kcoeffs}\\
\subjto \quad
  & \elnet \norm{\kcoeffs}_1 + (1-\elnet) \norm{\kcoeffs}_2^2 \leq 1,\\
  & \kcoeffs \vectgeq 0
\end{split}
\label{eq:elnet_LP}
\end{equation}
with $\normK \vectgeq 0$, $\normK \neq \vectzeros_\numk$ and $\elnet \in [0,1]$.
As shown in \sectionref{sec:lower_bound_MKL}, a solution
to this problem provides a lower bound on the optimal value
of the original MKL cost function \eqref{eq:GMKL}.

\subsection{Algorithm}
\label{sec:elnet_lp_algo}

In the special case $\elnet = 1$,
the (possibly nonunique) straightforward solution to the problem is
the vector $\vectbasis_\kk$,
where $\kk$ is such that $\normK_\kk = \max_i \normK_i$.
When $\elnet < 1$, simple algebra shows that points in $\Rnneg^\numk$
satisfy the elastic-net constraint if and only if they
also belong to the hyper-sphere
with centre $\centreSph$ and radius $\radiusSph$,
where
\begin{align}
\gapSph &= \elnet / (2 - 2\elnet),\\
\centreSph &= -\gapSph \, \vectones_\numk,\\
\radiusSph &= \sqrt{\numk \, \gapSph^2 + 2\gapSph + 1}.
\end{align}
Therefore, the problem \eqref{eq:elnet_LP}
is equivalent to:
\begin{equation}
\maximize\limits_{\substack{
\kcoeffs\in\Rnneg^\numk,\\
\norm{\kcoeffs-\centreSph}_2^2 \leq \radiusSph^2\\
}}
\; \scal{\normK}{\kcoeffs}.
\label{eq:projection_sphere}
\end{equation}
Let us now consider
the point $\tgtSph$:
\begin{equation}
\tgtSph = \radiusSph \, \normK/\norm{\normK}_2 + \centreSph,
\end{equation}
which is the point of the hyper-sphere which is farthest away
in the direction of $\normK$.
If this point is also in $\Rnneg^\numk$,
then $\kcoeffssollb=\tgtSph$ is trivially a solution for the
optimization problem \eqref{eq:projection_sphere}.
If this is not the case, the important property that
$\tgtSph_\kk < 0 \;\Rightarrow\; \kcoeffssollb_\kk = 0$
(of which a proof is provided in \sectionref{sec:elnet_lp_conv})
suggests a method to incrementally prune away coordinate
directions that are guaranteed to be zero in the
optimal solution $\kcoeffssollb$.
At each iteration $m$, the algorithm keeps track of
the set $\setzero\curr$ of indices for which
it has already been established that the corresponding
element of $\kcoeffssollb$ is null,
\ie $\kk \in \setzero\curr \Rightarrow \kcoeffssollb_\kk = 0$.
The set
$\setzero$ is initialized to the empty set $\emptyset$
at the beginning of the algorithm and grows
monotonically at each iteration.
We denote as $\numel{\setzero}$ the cardinality of $\setzero$,
as $\complem{\setzero}$ its complement and as
$\normK_{\complem{\setzero}}$ the projection of $\normK$
on the $(\numk{-}\numel{\setzero})$\nbd dimensional
subspace spanned by coordinate directions
corresponding to indices in $\complem{\setzero}$.
The algorithm generates the next iterate $\tgtSph\curr$ according to:
\begin{equation}
\tgtSph\curr_\kk =
\begin{cases}
\radiusSph\curr \, \normK_\kk / \norm{\normK_{\complem{\setzero}\curr}}_2 - \gapSph,
   & \text{if $\kk \in \complem{\setzero}$,} \\
0  & \text{if $\kk \in \setzero$.} \\
\end{cases}
\label{eq:elnet_lp_nextqm}
\end{equation}
This is the point of the $\numel{\complem{\setzero}\curr}$\nbd dimensional disc
of radius
$\radiusSph\curr = \sqrt{\numel{\complem{\setzero}\curr}
    \, \gapSph^2 + 2\gapSph + 1}$
and centre
$\centreSph_{\complem{\setzero}\curr}$
which is farthest away in the direction of $\normK$.
If any of the elements of $\tgtSph\curr$ is negative,
their indices are added to $\setzero$ and the algorithm
starts a new iteration, otherwise the algorithm ends and the
last iterate is returned as the solution $\kcoeffssollb$
to the elastic-net constrained linear program \eqref{eq:elnet_LP}.
The detailed algorithm is reported in \algoref{algo:elnet_LP}.

  \begin{small}
  \begin{algorithm2e} 
  \Fn{\FSolveElNetLP{$\normK$}}{
    \tcp{Initialization}
    $\setzero \leftarrow \emptyset$\;
    $\gapSph \leftarrow \elnet / (2 - 2\elnet)$\;
    \Do{$\setneg \neq \emptyset$}{
      \tcp{Main loop}
      $\radiusSph \leftarrow \sqrt{\numel{\complem{\setzero}}\,\gapSph^2 + 2\gapSph + 1}$\;
      $\tgtSph_{\complem{\setzero}} \leftarrow
         \radiusSph\, \normK_{\complem{\setzero}} / \norm{\normK_{\complem{\setzero}}}_2 - \gapSph$\;
      $\setneg \leftarrow \{\kk \,|\, \tgtSph_\kk < 0\}$\;
      $\tgtSph_{\setneg} \leftarrow 0$\;
      $\setzero \leftarrow \setzero \cup \setneg$\;
    }
    \KwRet{$\tgtSph$}\;
  }
  \caption{Solve elastic-net constrained linear program.}\label{algo:elnet_LP}
  \end{algorithm2e}
  \end{small}

\subsection{Convergence analysis}
\label{sec:elnet_lp_conv}

The fact that the greedy algorithm presented
in \sectionref{sec:elnet_lp_algo} finds the global
solution in a finite number of iterations stems
from the property that if the algorithm
produces an iterate with a negative component,
the corresponding element of the solution must be zero:
\begin{equation}
\tgtSph\curr_\kk < 0 \;\Rightarrow\; \kcoeffssollb_\kk = 0, \quad \forall \, \kk,m.
\label{eq:elnet_LP_property}
\end{equation}
Aiming for a contradiction, let us assume that
$\kcoeffs$, with $\kcoeffs_\kk > 0$, is a solution to
\eqref{eq:elnet_LP}
and that at some point the algorithm produces the iterate
$\tgtSph\curr$ with $\tgtSph\curr_\kk < 0$.
For conciseness, we denote
$\tgtSph\curr$ simply as $\tgtSph$,
$\setzero\curr$ as $\setzero$,
$\radiusSph\curr$ as $\radiusSphcurr$,
and
$\normK_{\complem{\setzero}\curr}$ as $\normKcurr$,
within this section.
From \eqref{eq:elnet_lp_nextqm}, it follows that $\tgtSph_\kk < 0$ implies
$\normKcurr_\kk < \norm{\normKcurr} \, \gapSph / \radiusSphcurr$.
Let us consider the point
\begin{equation}
\kcoeffs' = \kcoeffs - \epsilon\,\vectbasis_\kk + \delta\,\normKcurr, \quad
\text{with } 0 < \epsilon \leq \kcoeffs_\kk
\text{ and } \delta = \frac{\gapSph\,\epsilon}{\radiusSphcurr\,\norm{\normKcurr}},
\end{equation}
and show that it satisfies the constraints of
\eqref{eq:projection_sphere}.
Because $\epsilon \leq \kcoeffs_\kk$, $\delta > 0$,
and $\normKcurr \vectgeq 0$, then
$\kcoeffs \vectgeq 0 \,\Rightarrow\, \kcoeffs' \vectgeq 0$.
It is now sufficient to show that
$\norm{\kcoeffs - \centreSph}^2 \leq \radiusSph
   \,\Rightarrow\, 
 \norm{\kcoeffs' - \centreSph}^2 \leq \radiusSph$:
\begin{equation}
\begin{split}
\norm{\kcoeffs' - \centreSph}^2
  &= \norm{\kcoeffs - \centreSph}^2
    + \norm{\delta\,\normKcurr - \epsilon\,\vectbasis_\kk}^2
    + 2\,\scal{(\delta\,\normKcurr - \epsilon\,\vectbasis_\kk)}{(\kcoeffs - \centreSph)}
    \\
  &= \norm{\kcoeffs - \centreSph}^2
    + \frac{\gapSph^2 \epsilon^2}%
           {\radiusSphcurr^2} + \epsilon^2
    - 2\,\delta\,\epsilon\,\normKcurr_\kk
    + 2 \, \frac{\gapSph\,\epsilon}{\radiusSphcurr\,\norm{\normKcurr}}\,
        \scal{\normKcurr}{(\kcoeffs_{\complem{\setzero}}
                           - \centreSph_{\complem{\setzero}})}
    - 2\,\epsilon\,\kcoeffs_\kk - 2\,\gapSph\,\epsilon
    \\
  &\leq \norm{\kcoeffs - \centreSph}^2
    + (2\,\epsilon^2 - 2\,\epsilon\,\kcoeffs_\kk)
    - 2\,\delta\,\epsilon\,\normKcurr_\kk
    + 2\,\gapSph\,\epsilon\,\left( \frac{\norm{\normKcurr}\,
         \norm{\kcoeffs_{\complem{\setzero}} - \centreSph_{\complem{\setzero}}}}%
         {\norm{\normKcurr}\,\radiusSphcurr} - 1 \right)
    \\
  &\leq \norm{\kcoeffs - \centreSph}^2
    + 2\,\gapSph\,\epsilon\,\left( \frac{
         \sqrt{\norm{\kcoeffs - \centreSph}^2 - \numel{\setzero}\,\gapSph^2 }}%
         {\radiusSphcurr} - 1 \right)
   \leq \norm{\kcoeffs - \centreSph}^2
\end{split}
\end{equation}
This proves that $\kcoeffs'$ is a feasible point for
\eqref{eq:projection_sphere}.
Because
$\scal{\normK}{\kcoeffs'} = \scal{\normK}{\kcoeffs}
  - \epsilon\,\normK_\kk + \delta\,\scal{\normK}{\normKcurr}
=  \scal{\normK}{\kcoeffs} - \epsilon\,\normKcurr_\kk
  + \frac{\gapSph\,\epsilon}{\radiusSphcurr\,\norm{\normKcurr}} \norm{\normKcurr}^2
> \scal{\normK}{\kcoeffs}$,
the feasible point $\kcoeffs'$ improves over $\kcoeffs$,
which therefore cannot be a solution.
This contradiction proves \eqref{eq:elnet_LP_property}.

\section{Experimental Results}
\label{sec:exp_results}


\subsection{Elastic-net constrained MKL}

This section presents the results of a comparison between
the efficiency of \algoref{algo:outer}
in solving the elastic-net constrained MKL
problem an that of the state-of-the-art
cutting-plane method by \citet{yang_efficient_2011}
as implemented in the GMKL toolbox
available from the authors' homepage.
Note that the two algorithms solve the same convex
optimization problem and, thus, converge to the
same solution (or solution set).
For this reason, only the training time is reported here,
since the number of kernels selected and the
prediction accuracy on the test set are very similar.
The analysis was performed measuring
the running time
on a computer with
Intel$^{\circledR}$ Core(TM) i7-5500U CPU running
Matlab$^{\circledR}$ 2015b 64bit on a single core
under Ubuntu 16.04.

The algorithms' efficiency was assessed both on synthetic and on
real-world data.
The synthetic datasets, \dataset{Toy1} and \dataset{Toy2},
were obtained as described in Section VI.B of \citep{yang_efficient_2011}.
The real-world data included the
\dataset{Breast}, \dataset{Heart}, \dataset{Ionosphere}, \dataset{Liver}, \dataset{Pima}, \dataset{Sonar}, \dataset{Wdbc}, and \dataset{Wpbc} datasets
from the UCI repository \citep{Dua_2019},
that have been previously used for similar purposes
\citep{rakotomamonjy_simplemkl_2008,xu_simple_2010,yang_efficient_2011}.

In all experiments, binary classifiers
were trained using $60\%$ randomly selected
examples and then tested on the remaining data.
This procedure was repeated $5$ times for each dataset
and configuration.
All feature vectors were normalized such that each feature
had zero mean and unit variance on the training set.
As in previous research
\citep{rakotomamonjy_simplemkl_2008,xu_simple_2010,yang_efficient_2011}
the base Gram matrices were built using RBF kernels
with widths $\{2^{-3}, 2^{-2}, \dots, 2^6\}$
and polynomial kernels of degrees $\{1, 2, 3\}$,
all computed on each single feature and on the whole feature vector.
Gram matrices were pre-computed and normalized
to have unit trace.

All MKL models used the same SVM solver,
originally developed by \citet{SVM-KMToolbox},
with default settings.
Since the focus of this comparison is on the training time
rather than classification accuracy, the regularization
parameter of the SVM was set to a constant value $C=100$
rather than tuned using a nested cross-validation loop.
This constant value yielded accuracies and
number of active kernels similar to those previously reported
by others \citep{xu_simple_2010,yang_efficient_2011}
on the same data.
As back-end solver for the cutting-plane method, the tests included both the
open-source CVX solver \citep{cvx}
and the
commercial Mosek solver \citep{mosek} using conic optimization
as suggested in the GMKL toolbox.
The relative gap between the
cost function \eqref{eq:GMKL} of the MKL problem and its lower bound
was used as stopping criterion for the two-step block
coordinate descent.
For the novel algorithm introduced here,
the lower bound was computed as described
in \sectionref{sec:lower_bound_MKL},
while for the cutting-plane method
equation (22) of \citep{yang_efficient_2011} was used.
The algorithms were also terminated if they failed
to reach the desired accuracy $\epsilon_{\mathrm{MKL}}$
within 500 iterations.

\begin{figure}[tb]
    \centering
    \includegraphics[width=\textwidth]{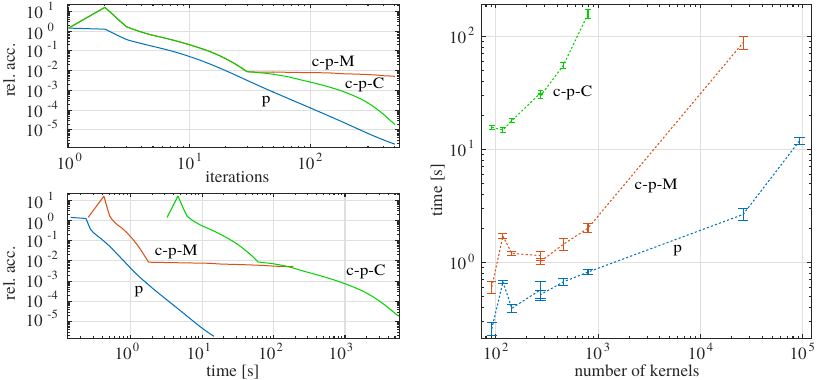}
    \caption{On the left, Relative accuracy of the different
        algorithms
        (p: proposed; c-p-M: cutting-plane with Mosek solver;
        c-p-C: cutting-plane with CVX solver)
        as a function of the number of iterations
        (top) or the computation time (bottom) on the
        \dataset{Ionosphere} dataset.
        On the right, computation time required to
        achieve a relative accuracy $10^{-2}$ on the different
        datasets, plotted as a function of the number of kernels
        (missing points are due to algorithms exceeding time or
        memory limits).
    }\label{fig:iterations}
\end{figure}
\figureref{fig:iterations}-left compares the accuracy
of each algorithm measured as the relative gap
between \eqref{eq:GMKL} and \eqref{eq:elnet_LP_for_MKL}.
As the figure shows, the algorithm presented in this paper
is significantly faster than the cutting-plane method
using either the open source CVX toolbox or the commercial
MOSEK solver.
From \figureref{fig:iterations}-left, we can see that
the cutting-plane method with the MOSEK solver struggles
to improve the accuracy past a certain limit while the new algorithm
steadily improves to much smaller values.

Even discounting for this issue, the proposed algorithm
represents a significant improvement over the previous method,
as shown in \figureref{fig:iterations}-right where
a relative gap $10^{-2}$ was used as stopping criterion.
The improvement is even larger when requiring
a relative gap $10^{-3}$ or smaller (not reported),
which is consistent with the results of \figureref{fig:iterations}-left.
Finally, it is worth mentioning that the space complexity
of the proposed algorithm scales linearly with the number of kernels
while it scales quadratically in the case of the cutting-plane method.

\section{Discussion and Conclusions}
\label{sec:conclusions}

This paper presents a novel algorithm for the
solution of elastic-net constrained multiple kernel learning problems.
Analysis of the computational cost of the
algorithm shows that it compares very favourably to existing
alternative approaches.

While solving the main MKL problem, efficient algorithms were
also devised for more general optimisation problems,
namely elastic-net constrained weighted sums of reciprocals
and elastic-net constrained linear programs.
These algorithms have general applicability also outside the
domain of multiple kernel learning.

Finally, because the proposed algorithm does
not depend on external libraries,
it has a wide applicability and can be readily
included in existing open-source machine learning libraries.

\appendix

\section{Proofs of theorems}
The proofs of the theorems given in the text
are reported in this appendix in the form of structured proofs as
advocated by Leslie Lamport \citeyearpar{lamport_how_2012}.
Each assertion follows from previously stated facts,
which are explicitly named to tell the reader
exactly which ones are being used at each step.

\subsection{Proofs of \theoremref{th:pseudoconvexity} and \corollaryref{th:pseudoconvexitycoroll}}

\label{sec:proof_pseudoconvexity}

\pseudoconvexity*

\begin{proof}\hspace{1ex}
\begin{lsp}[series=lsp-pseudoconv]
  \item To show that the differentiable function
  $\fnPr: \setA \rightarrow \Rpos$ defined in an open convex set
  is pseudoconvex, it suffices to assume for the
    remaining of this proof that: \label{th:assump_pseudo}
  \begin{lsp}
    \item $\xint,\xnext \in \setA$, \label{th:assump_pts}
    \item $\fnPr(\xnext) < \fnPr(\xint)$, \label{th:assump_fn}
  \end{lsp}
  and prove that
  $\scal{\grad{\fnPr(\xint)}}{(\xnext-\xint)} < 0$.
  \lspproof By the definition of pseudoconvex function
    \citep[definition 3.2.1]{cambini_generalized_2008}.

  \item $\forall\, \x \in \setA:
        \scal{\grad{\fnSr(\x)}}{\x} = -\fnSr(\x)$. \label{th:fn_grad_Sr}
  \lspproof By differentiating $\fnSr(c\x) = \fnSr(\x)/c$
  w.r.t.\ $c$ and evaluating it for $c=1$.

  \item $\forall\, \x \in \setA:
      \scal{\grad{\fnSc(\x)}}{\x} = \fnSc(\x)$. \label{th:fn_grad_Sc}
  \lspproof By differentiating $\fnSc(c\x) = c\fnSc(\x)$
    w.r.t.\ $c$ and evaluating it for $c=1$.

  \item $\forall\, \x \in \setA:
      \scal{\grad{\fnPr(\x)}}{\x} =
      \fnSr(\x)  \scal{\grad{\fnSc(\x)}}{\x} +
    \fnSc(\x)  \scal{\grad{\fnSr(\x)}}{\x} = 0$.\label{th:fn_grad_zero}
  \lspproof Follows directly from \ref{th:fn_grad_Sr} and \ref{th:fn_grad_Sc}.

  \item Given $\xnext$ and $\xint$ as in \ref{th:assump_pts},
    $\exists \, c \in \Rpos$ such that the point
    $\xnext' = c\xnext$ satisfies $\fnPr(\xnext') = \fnPr(\xnext)$ and $\fnSc(\xnext') = \fnSc(\xint)$.
    \label{th:scaledpoint}
  \lspproof
    For any positive $c$ the corresponding $\xnext'$ is in $\setA$ (because
     $\setA$ is a cone) and satisfies the first condition:
             $\fnPr(\xnext') = \fnSc(c\xnext)\,\fnSr(c\xnext)
              = c\fnSc(\xnext)\,\fnSr(\xnext)/c =  \fnPr(\xnext)$.
    We choose $c = \fnSc(\xint)/\fnSc(\xnext)$ which also satisfies the second condition:
    $\fnSc(\xnext') = \fnSc(\xint)/\fnSc(\xnext) \, \fnSc(\xnext) = \fnSc(\xint)$.

  \item $\forall\, \x,\x' \in \setA:
      \scal{\grad{\fnSc(\x)}}{\x'} \leq \fnSc(\x')$.
    \label{th:fnSc_leq}
  \lspproof
    The first-order conditions for convexity \citep[ch 3.1.3]{boyd_convex_2009}
    imply $\fnSc(\x') \geq \fnSc(\x) + \scal{\grad{\fnSc(\x)}}{(\x'-\x)}$.
    Substituting \ref{th:fn_grad_Sc} and rearranging yields \ref{th:fnSc_leq}.

  \item $\forall\, \x,\x' \in \setA:
      \scal{\grad{\fnSr(\x)}}{\x'} \leq \fnSr(\x') - 2 \, \fnSr(\x)$.
    \label{th:fnSr_leq}
  \lspproof
    The first-order conditions for convexity 
    imply $\fnSr(\x') \geq \fnSr(\x) + \scal{\grad{\fnSr(\x)}}{(\x'-\x)}$.
    Substituting \ref{th:fn_grad_Sr} and rearranging yields \ref{th:fnSr_leq}.

  \item $\scal{\grad{\fnPr(\xint)}}{\xnext'} < 0$. \label{th:gradPr_leq0}
    \lspproof By \ref{th:fnSc_leq}, \ref{th:fnSr_leq}, \ref{th:scaledpoint}
     and \ref{th:assump_fn}, we have:
      \begin{align*}
      \scal{\grad{\fnPr(\xint)}}{\xnext'} &=
            \fnSr(\xint) \, \scal{\grad{\fnSc(\xint)}}{\xnext'} +
            \fnSc(\xint) \, \scal{\grad{\fnSr(\xint)}}{\xnext'} \\
            &\leq \fnSr(\xint) \, \fnSc(\xnext') +
                  [\fnSc(\xint) \, \fnSr(\xnext') - 2 \, \fnSc(\xint) \, \fnSr(\xint)] \\
            &= \fnPr(\xint) + \fnPr(\xnext') - 2 \, \fnPr(\xint) \\
            &= \fnPr(\xnext') - \fnPr(\xint) = \fnPr(\xnext) - \fnPr(\xint) < 0.
      \end{align*}

    \item \lspqed
      \lspproof By \ref{th:gradPr_leq0}, \ref{th:scaledpoint} and \ref{th:fn_grad_zero},
        we have:
        $$
        c\scal{\grad{\fnPr(\xint)}}{\xnext} < 0 \;\Rightarrow\;
                    \scal{\grad{\fnPr(\xint)}}{\xnext} =
                    \scal{\grad{\fnPr(\xint)}}{(\xnext-\xint)} < 0.
        $$
        By \ref{th:assump_pseudo}, the latter proves the theorem.

\end{lsp}
\end{proof}

\pseudoconvexitycoroll*

\begin{proof}\hspace{1ex}
\begin{lsp}[resume*=lsp-pseudoconv,widest=19]
  \item $\grad{\fnSc(\xsol)} = -\grad{\fnSr(\xsol)} \;\Rightarrow\;
       \fnSc(\xsol) = \fnSr(\xsol)$.\label{th:same_grad_same_val}
  \lspproof Follows immediately from the statements \ref{th:fn_grad_Sr}
  and \ref{th:fn_grad_Sc} of the proof of \theoremref{th:pseudoconvexity}.

  \item $\xsol$ is a critical point for $\fnPr$.
  \lspproof From the condition $\grad{\fnSc(\xsol)} = -\grad{\fnSr(\xsol)}$
      and from statement \ref{th:same_grad_same_val}:
      $\grad \fnPr(\xsol) = \fnSr(\xsol)\,\grad \fnSc(\xsol) + \fnSc(\xsol)\,\grad \fnSr(\xsol) = 0$.

  \item $\xsol$ is a global minimum of $\fnPr$.\label{th:critical_point}
  \lspproof Because $\xsol$ is a critical point (statement \ref{th:critical_point})
      of a pseudoconvex function (\theoremref{th:pseudoconvexity}),
      it is also a global minimum
      \citep[theorem 3.2.5]{cambini_generalized_2008}.

  \item If at least one of $\fnSc$ or $\fnSr$ is strictly convex, then
      $\xsol$ is unique.\label{th:xsol_unique}
  \lspproof Aiming for a contradiction, let us assume that there
      is a point $\hat\x \in \setA \setminus \{\xsol\}$ such that
      $\grad{\fnSc(\hat\x)} = -\grad{\fnSr(\hat\x)}$.
      By using the same reasoning as in \ref{th:same_grad_same_val},
      this implies $\fnSc(\hat\x) = \fnSr(\hat\x)$.
      Without loss of generality, let us assume that
      $\fnSc(\xsol) \geq \fnSc(\hat\x)$ and that $\fnSc$ is strictly convex.
      From the first-order conditions for (strict) convexity,
      we obtain:
      \begin{alignat}{2}
        \fnSc(\hat\x) &>    \fnSc(\xsol) + \scal{\grad{\fnSc(\xsol)}}{(\hat\x-\xsol)}
        &\quad\Rightarrow\quad& \scal{\grad{\fnSc(\xsol)}}{(\hat\x-\xsol)} < 0,
        \\
        \fnSr(\hat\x) &\geq \fnSr(\xsol) + \scal{\grad{\fnSr(\xsol)}}{(\hat\x-\xsol)}
        &\quad\Rightarrow\quad& \scal{\grad{\fnSc(\xsol)}}{(\hat\x-\xsol)} \geq 0,
      \end{alignat}
      which is obviously a contradiction.

  \item \lspqed
  \lspproof From \ref{th:same_grad_same_val}, \ref{th:critical_point},
      \ref{th:xsol_unique}, and 
      the definitions of $\fnSc$, $\fnSr$, and $\fnPr$
      in the statement of \theoremref{th:pseudoconvexity}.

\end{lsp}
\end{proof}

\subsection{Proof of \theoremref{th:boundsfnSc}}
\label{sec:proof_boundsfnSc}

\boundsfnSc*

\begin{proof}\hspace{1ex}
\begin{lsp}
  \item $\forall\, \xint \in \mathbb{R}^n:
      \scal{\grad{\fnSc(\xint)}}{\xint} = \fnSc(\xint)$. \label{th:fn_grad_Sc2}
  \lspproof Since $\fnSc$ is a norm, $\fnSc(c\xint) =  \abs{c} \fnSc(\xint)$. By differentiating both sides w.r.t.\ $c$ and evaluating it for $c=1$, we
  obtain the statement \ref{th:fn_grad_Sc2}.

  \item $\forall\, \xint,\x \in \mathbb{R}^n:
      \scal{\grad{\fnSc(\xint)}}{\x} \leq \fnSc(\x)$.
    \label{th:fnSc_leq_grad}
  \lspproof
    The first-order conditions for convexity \citep[ch 3.1.3]{boyd_convex_2009}
    imply $\fnSc(\x) \geq \fnSc(\xint) + \scal{\grad{\fnSc(\xint)}}{(\x-\xint)}$.
    Substituting \ref{th:fn_grad_Sc2} and rearranging yields \ref{th:fnSc_leq_grad}.

  \item Define $r: \mathbb{R}^n \rightarrow \Rnneg$ as
        $r(\xint) = \sqrt{d_1^2 \norm{\xint}_1^2 + d_2^2 \norm{\xint}_2^2}$.

  \item $\displaystyle \forall\, \xint,\x \in \mathbb{R}^n:
    \sqrt{ \frac{ d_1^2 \norm{\x}_1^2 + d_2^2 \norm{\x}_2^2 }{ \norm{\x}_1^2 } }
    \leq
    \frac{1}{2} \biggl[\frac{r(\xint)}{\norm{\xint}_1}
      + \frac{d_1^2 \norm{\x}_1^2 + d_2^2 \norm{\x}_2^2}{\norm{\x}_1^2}
         \frac{\norm{\xint}_1}{r(\xint)} \biggr]$. \label{th:bound_sqrt}
  \lspproof
    From the inequality $\sqrt{\xnext} \leq \frac{1}{2}
        [\sqrt{\xnext_0} + \xnext / \sqrt{\xnext_0}]$, which in turn results from
        the concavity of the square root function.

  \item $\displaystyle \forall\, \xint,\x \in \mathbb{R}^n:
    \fnSc^2(\x) \leq
    \fnSc(\xint) \biggl[
      \frac{d_0}{\norm{\xint}_1} \norm{\x}_1^2
      + \frac{d_1}{r(\xint)} \norm{\x}_1^2
      + \frac{d_2}{r(\xint)} \norm{\x}_2^2
      \biggr]$.\label{th:boundsfnSc_ineq1}
  \lspproof
     Follows from writing out the lhs explicitly using the definition of $\fnSc$
     and then exploiting the statement in \ref{th:bound_sqrt}.

  \item $\displaystyle \forall\, \xint,\x \in \mathbb{R}^n:
      \frac{d_0}{\norm{\xint}_1} \norm{\x}_1^2
      + \frac{d_1}{r(\xint)} \norm{\x}_1^2
      + \frac{d_2}{r(\xint)} \norm{\x}_2^2
      \leq
            \suml_i \frac{d_0}{\abs{\xint_i}} \x_i^2
            + \suml_i \frac{d_1 \norm{\xint}_1}{r(\xint)\abs{\xint_i}} \x_i^2
            + \suml_i \frac{d_2}{r(\xint)} \x_i^2$.\label{th:boundsfnSc_ineq2}
  \lspproof
     The last term of each side of the inequality is identical.
     Applying Radon's inequality it is easy to show that the each one
     of the first two terms of the lhs is bounded by the corresponding
     term in the rhs.

  \item $\displaystyle \forall\, \xint,\x \in \mathbb{R}^n:
      \fnSc^2(\x) \leq \x\TR \Mdiag_{\xint} \, \x$.\label{th:fnSc_leq_quad}
  \lspproof
     Follows from combining \ref{th:boundsfnSc_ineq1} and \ref{th:boundsfnSc_ineq2}, then
     using the definition of $\Mdiag_{\xint}$.

  \item \lspqed
  \lspproof
      Combining \ref{th:fnSc_leq_grad} and \ref{th:fnSc_leq_quad}
      proves the theorem.
\end{lsp}
\end{proof}


\begin{thebibliography}{18}
\providecommand{\natexlab}[1]{#1}
\providecommand{\url}[1]{\texttt{#1}}
\expandafter\ifx\csname urlstyle\endcsname\relax
  \providecommand{\doi}[1]{doi: #1}\else
  \providecommand{\doi}{doi: \begingroup \urlstyle{rm}\Url}\fi

\bibitem[Bertsekas(1999)]{bertsekas_nonlinear_1999}
Dimitri~P. Bertsekas.
\newblock \emph{Nonlinear {Programming}}.
\newblock Athena Scientific, Belmont, Mass, 2nd edition edition, September
  1999.
\newblock ISBN 978-1-886-52900-7.

\bibitem[Boyd and Vandenberghe(2009)]{boyd_convex_2009}
Stephen Boyd and Lieven Vandenberghe.
\newblock \emph{Convex optimization}.
\newblock Cambridge university press, 2009.
\newblock ISBN 978-0-521-83378-3.

\bibitem[Cambini and Martein(2008)]{cambini_generalized_2008}
Alberto Cambini and Laura Martein.
\newblock \emph{Generalized convexity and optimization: Theory and
  applications}, volume 616.
\newblock Springer, 2008.
\newblock ISBN 978-3-540-70875-9.

\bibitem[Canu et~al.(2005)Canu, Grandvalet, Guigue, and
  Rakotomamonjy]{SVM-KMToolbox}
S.~Canu, Y.~Grandvalet, V.~Guigue, and A.~Rakotomamonjy.
\newblock {SVM} and kernel methods {M}atlab toolbox.
\newblock Perception Syst\`emes et Information, INSA de Rouen, Rouen, France,
  2005.

\bibitem[Citi(2015)]{citi_elastic-net-MM_2015}
Luca Citi.
\newblock Elastic-net constrained multiple kernel learning using a
  majorization-minimization approach.
\newblock In \emph{Proceedings of the 7th Computer Science and Electronic
  Engineering Conference (CEEC)}, pages 29--34, 2015.
\newblock \doi{10.1109/CEEC.2015.7332695}.

\bibitem[Dua and Graff(2019)]{Dua_2019}
Dheeru Dua and Casey Graff.
\newblock {UCI} machine learning repository, 2019.
\newblock URL \url{http://archive.ics.uci.edu/ml}.

\bibitem[Grant and Boyd(2015)]{cvx}
Michael Grant and Stephen Boyd.
\newblock {CVX}: {M}atlab software for disciplined convex programming, version
  2.1.
\newblock \url{http://cvxr.com/cvx}, 2015.

\bibitem[Lamport(2012)]{lamport_how_2012}
Leslie Lamport.
\newblock How to write a 21st century proof.
\newblock \emph{Journal of Fixed Point Theory and Applications}, 11\penalty0
  (1):\penalty0 43--63, 2012.
\newblock \doi{10.1007/s11784-012-0071-6}.

\bibitem[Luenberger and Ye(2008)]{luenberger_linear_2008}
David~G. Luenberger and Yinyu Ye.
\newblock \emph{Linear and Nonlinear Programming}.
\newblock Springer Science \& Business Media, June 2008.
\newblock ISBN 978-0-387-74503-9.

\bibitem[Meyer(1976)]{meyer_sufficient_1976}
Robert~R. Meyer.
\newblock Sufficient conditions for the convergence of monotonic mathematical
  programming algorithms.
\newblock \emph{Journal of Computer and System Sciences}, 12\penalty0
  (1):\penalty0 108--121, February 1976.
\newblock \doi{10.1016/S0022-0000(76)80021-9}.

\bibitem[{MOSEK ApS}(2015)]{mosek}
{MOSEK ApS}.
\newblock The {MOSEK} optimization toolbox for {M}atlab, version 7.1.0.33.
\newblock \url{https://www.mosek.com/products/mosek}, 2015.

\bibitem[Rakotomamonjy et~al.(2007)Rakotomamonjy, Bach, Canu, and
  Grandvalet]{rakotomamonjy_more_2007}
Alain Rakotomamonjy, Francis Bach, St{\'e}phane Canu, and Yves Grandvalet.
\newblock More efficiency in multiple kernel learning.
\newblock In \emph{Proceedings of the 24th international conference on Machine
  learning}, pages 775--782. {ACM}, 2007.

\bibitem[Rakotomamonjy et~al.(2008)Rakotomamonjy, Bach, Canu, and
  Grandvalet]{rakotomamonjy_simplemkl_2008}
Alain Rakotomamonjy, Francis Bach, St{\'e}phane Canu, and Yves Grandvalet.
\newblock {SimpleMKL}.
\newblock \emph{Journal of Machine Learning Research}, 9:\penalty0 2491--2521,
  2008.

\bibitem[Rockafellar(1970)]{rockafellar_convex_1970}
Ralph~T. Rockafellar.
\newblock \emph{Convex Analysis}.
\newblock Princeton University Press, 1970.
\newblock ISBN 0-691-08069-0.

\bibitem[Sun et~al.(2013)Sun, Jiao, Liu, Wang, and Feng]{sun_selective_2013}
Tao Sun, Licheng Jiao, Fang Liu, Shuang Wang, and Jie Feng.
\newblock Selective multiple kernel learning for classification with ensemble
  strategy.
\newblock \emph{Pattern Recognition}, 46\penalty0 (11):\penalty0 3081--3090,
  November 2013.
\newblock \doi{10.1016/j.patcog.2013.04.003}.

\bibitem[Xu et~al.(2010)Xu, Jin, Yang, King, and Lyu]{xu_simple_2010}
Zenglin Xu, Rong Jin, Haiqin Yang, Irwin King, and Michael~R. Lyu.
\newblock Simple and efficient multiple kernel learning by group lasso.
\newblock In \emph{Proceedings of the 27th International Conference on Machine
  Learning ({ICML}-10)}, pages 1175--1182, 2010.

\bibitem[Yang et~al.(2011)Yang, Xu, Ye, King, and Lyu]{yang_efficient_2011}
Haiqin Yang, Zenglin Xu, Jieping Ye, I~King, and M~R Lyu.
\newblock Efficient sparse generalized multiple kernel learning.
\newblock \emph{{IEEE} Transactions on Neural Networks}, 22\penalty0
  (3):\penalty0 433--446, March 2011.
\newblock \doi{10.1109/TNN.2010.2103571}.

\bibitem[Zangwill(1969)]{zangwill_nonlinear_1969}
Willard~I. Zangwill.
\newblock \emph{Nonlinear programming: a unified approach}.
\newblock Prentice-Hall, 1969.
\newblock ISBN 978-0-136-23579-8.

\end{thebibliography}

\end{document}